\documentclass[10pt,twocolumn,letterpaper]{article}

\usepackage{cvpr}              
\definecolor{cvprblue}{rgb}{0.21,0.49,0.74}
\usepackage[pagebackref,breaklinks,colorlinks,allcolors=cvprblue]{hyperref}

\usepackage{multirow}
\usepackage{colortbl}
\usepackage{xcolor}

\usepackage{graphicx}
\usepackage{caption}
\usepackage{multicol}

\usepackage[accsupp]{axessibility}

\usepackage{amsmath} 
\usepackage{xspace} 

\newcommand{\modelname}{\textsc{LivingSwap}\xspace}
\newcommand{\dataset}{\textit{Face2Face}\xspace}
\newcommand{\benchmark}{\textit{CineFaceBench}\xspace}

\def\confName{CVPR}
\def\confYear{2026}

\title{Preserving Source Video Realism: High-Fidelity Face Swapping \\
for Cinematic Quality}

\author{
\begin{tabular}{ccccc}
    Zekai Luo$^{1}$ ~~ 
    Zongze Du$^{1}$ ~~
    Zhouhang Zhu$^{1}$ ~~
    Hao Zhong$^{1}$ ~~
    Muzhi Zhu$^{1}$ \\
    Wen Wang$^{1}$ ~~
    Yuling Xi$^{1}$ ~~
    Chenchen Jing$^{2}$ ~~
    Hao Chen$^{1,\dagger}$ ~~
    Chunhua Shen$^{1,2,3,\dagger}$
\end{tabular} \\[.25cm]
\small{$^1$Zhejiang University, 
State Key Laboratory of CAD \& CG  \qquad $^2$Zhejiang University of Technology \qquad $^3$Ant Group}
}

\begin{document}

\twocolumn[{
    \renewcommand\twocolumn[1][]{#1}
    \vspace{-12mm}
    \maketitle
    \vspace{-10mm}
    \begin{center}
        \captionsetup{type=figure}
        \includegraphics[width=1.0\linewidth]{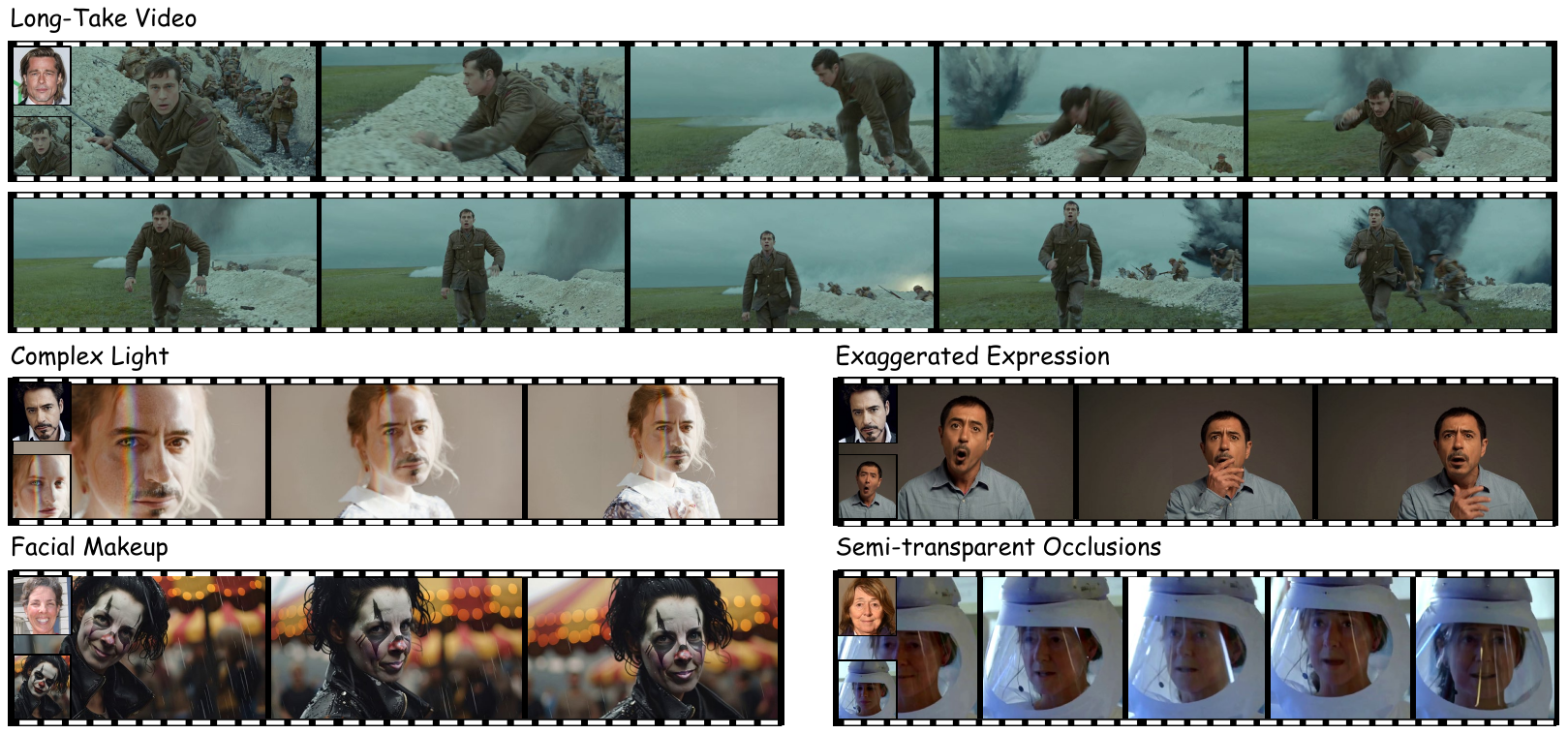}
        \caption{
        Qualitative results of our proposed video reference guided face swapping model, \modelname.
        Across challenging cinematic scenarios—including long-take shots, complex illumination, exaggerated expressions, heavy facial makeup, and semi-transparent occlusions—our method consistently preserves target identity and fine-grained attributes with high fidelity, while maintaining robust visual realism.
        }
        \label{fig:vis}
    \end{center}
}]


\newcommand\blfootnote[1]{%
  \begingroup
  \renewcommand\thefootnote{}\footnote{#1}%
  \addtocounter{footnote}{-1}%
  \endgroup
}

\blfootnote{$^\dagger$ Corresponding authors.}

\begin{abstract}


Video face swapping is crucial in film and entertainment production, where achieving high fidelity and temporal consistency over long and complex video sequences remains a significant challenge. 
Inspired by recent advances in reference-guided image editing, we explore whether rich visual attributes from source videos can be similarly leveraged to enhance both fidelity and temporal coherence in video face swapping.
Building on this insight, this work presents \modelname, the first video reference guided face swapping model. 
Our approach employs keyframes as conditioning signals to inject the target identity, enabling flexible and controllable editing. 
By combining keyframe conditioning with video reference guidance, the model performs temporal stitching to ensure stable identity preservation and high-fidelity reconstruction across long video sequences.
To address the scarcity of data for reference-guided training, we construct a paired face-swapping dataset, \dataset, and further reverse the data pairs to ensure reliable ground-truth supervision.
Extensive experiments demonstrate that our method achieves state-of-the-art results, seamlessly integrating the target identity with the source video’s expressions, lighting, and motion, while significantly reducing manual effort in production workflows.
Project webpage: 
\href{https://aim-uofa.github.io/LivingSwap}{\textcolor{cvprblue}{https://aim-uofa.github.io/LivingSwap}}

\end{abstract}    
\section{Introduction}
\label{sec:intro}

\begin{figure*}[t]
    \centering
    \includegraphics[width=0.85\linewidth]{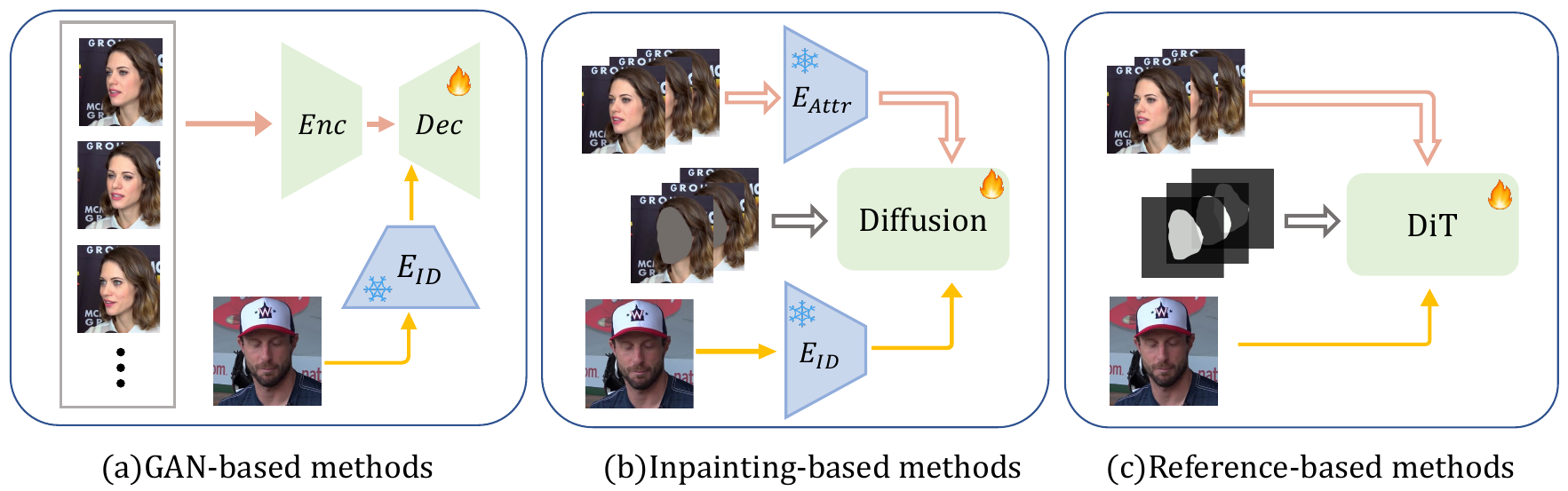}
    \vspace{-5pt}
    \caption{ 
    (a) GAN-based approaches process videos in a frame-by-frame manner, and therefore often struggle with realism and suffer from temporal inconsistency.
    (b) Inpainting-based methods focus on generating the facial region based on sparse conditions, which inevitably leads to a loss of fidelity and unnatural visual artifacts.
    (c) Recent reference-based generation methods enable faithful utilization of rich visual attributes contained in references and demonstrate remarkable capability in preserving them.
    }
    \vspace{-10pt}
    \label{fig:reference}
\end{figure*}

Video face swapping holds significant value in the film and entertainment industries. 
However, existing methods fall short of meeting the stringent demands of high-quality cinematic production. 
For instance, GAN-based methods~\citep{li2019faceshifter,deepfake,chen2020simswap,shiohara2023blendface,luo2025canonswap}, which typically operate in a frame-by-frame manner (see \cref{fig:reference}a), have made considerable progress in injecting the target identity. 
Yet they often struggle with realism and suffer from temporal artifacts—such as flickering and jitter—especially in long sequences.
Meanwhile, contemporary video diffusion models~\citep{zhao2023diffswap,han2024face,chen2024hifivfs,wang2025dynamicface,zhang2025vividface}, while achieving high visual quality and temporal consistency, often rely on intermediate representations from external encoders (e.g., facial landmarks or 3D faces), which inevitably lose information compared to raw source videos. 
This reliance makes it challenging to perfectly align the generated expressions, lighting, and subtle nuances with the source video, resulting in faces that may appear unnatural or lack lifelike vitality. 
Consequently, there is a critical need for a video face swapping model capable of directly leveraging the rich, detailed information from the source video's facial region.

Achieving a high degree of customization while preserving the integrity of the original content remains a fundamental challenge in generative media~\citep{yang2023object}. 
Methods based on DDIM inversion~\citep{ju2023direct,qi2023fatezero,geyer2023tokenflow} or Score Distillation Sampling (SDS)~\citep{poole2022dreamfusion,hertz2023delta} often struggle to strike an optimal balance between editability and fidelity. 
In the field of video editing, a common strategy involves combining inpainting with structural guidance such as depth or keypoints~\citep{jiang2025vace,hu2025animate,tu2025videoanydoor} (see \cref{fig:reference}b). 
However, such approaches inherently discard the original pixel information within the edited region, leading to a noticeable loss of fidelity in details.

Recently, reference guided generation has demonstrated remarkable breakthroughs in image editing, successfully reconciling editing flexibility with high-fidelity reconstruction~\citep{labs2025flux,deng2025emerging,wu2025qwen}.
This approach directly guides the model using the reference images, enabling the faithful utilization of rich visual attributes contained in the references.
Nevertheless, adapting these techniques to video face swapping presents unique challenges: 
(1) the difficulty of injecting a stable and consistent identity condition throughout long and complex video sequences; 
and (2) the scarcity of paired training data for reference-guided video face swapping task.


In this work, we address these challenges by introducing \modelname, the first video editing model for face swapping that directly references the source video's details. 
To facilitate this, we decompose the challenging task of long-video face swapping into a highly controllable pipeline comprising keyframe identity injection, video reference completion, and temporal stitching.
This pipeline not only enables flexible identity guidance using high-quality image swapping results, but also mitigates the accumulation of errors that typically arise in long videos.Furthermore, we construct \dataset, the first-of-its-kind dataset specifically curated for video reference-guided face swapping. To ensure reliable ground-truth supervision, we reverse each data pair by using the generated results as inputs and the original data as the ground truth.


To further validate the effectiveness of \modelname, we collect a set of cinematic video clips featuring a wide range of challenging conditions and construct the \benchmark.
Benefiting from its highly controllable pipeline and superior generation quality, \modelname seamlessly integrates the target identity with the high-definition details of the source video, faithfully preserving original expressions, lighting conditions, and other key facial attributes.



Our contributions are as follows:
\begin{itemize}
    \item We introduce \modelname, a stable video face-swapping solution with a controllable pipeline that reduces the need for frame-by-frame human editing by a factor of 40, making it particularly well-suited for the professional film and television industry.
    \item We introduce \dataset, a paired dataset designed to address the scarcity of training data for video reference–guided face swapping.
    Moreover, by reverse data pairs and leveraging strong priors from pretrained video models, our approach is able to surpass the limitations of the dataset itself and achieve superior performance.
    \item We propose \benchmark, a cinematic scenarios benchmark that facilitates reliable comparison of video face-swapping models tailored to industrial scenarios.
\end{itemize}

\section{Related Work}

\label{sec:relat}
    
\textbf{Video face swapping.}
The task of video face swapping is to replace the identity in a video while preserving attributes such as pose, expression, illumination, and background. 
GAN-based approaches~\citep{li2019faceshifter,deepfake,chen2020simswap,shiohara2023blendface,luo2025canonswap,tzaban2022stitch,jiang2024identity}, which typically process videos frame-by-frame, have made notable progress in injecting target identity through encoder–decoder pipelines. 
However, they often suffer from temporal inconsistencies—such as flickering and jitter—especially in long sequences.
Recently, video diffusion models are used in video face swapping.  
Diffusion-based methods  ~\citep{zhao2023diffswap,han2024face,chen2024hifivfs,wang2025dynamicface} demonstrate stronger generative power and achieve higher visual quality and temporal consistency.   
They treat face swapping as inpainting by masking the original face and regenerating it conditioned on background frames and auxiliary attributes~\citep{chen2024hifivfs,wang2025dynamicface}. 
This often leads to the loss of fine-grained details and introduces inconsistencies with the model’s pretrained priors, thereby degrading generation quality.  
In this work, we tackle these challenges by directly leveraging detailed source video references for face swapping, combined with a carefully curated dataset \dataset and a reversing data strategy to provide high-fidelity supervision.

\textbf{Diffusion-based Video Editing.}
With the rapid development of diffusion models, a wide range of customization and editing techniques~\cite{ruiz2023dreambooth,cao2023masactrl,yang2024lora,zhu2024unleashing,zhao2025local,zhao20253d,yang2025any} have emerged. 
From a methodological perspective, these approaches can be broadly categorized into inversion-based, inpainting-based, and reference-guided methods.
Inversion-based methods~\citep{ju2023direct,qi2023fatezero,geyer2023tokenflow,poole2022dreamfusion,hertz2023delta} reconstruct the original video trajectory in the diffusion process to enable editing, but they often struggle to balance editability and fidelity.
Inpainting-based approaches~\citep{jiang2025vace,hu2025animate,tu2025videoanydoor} edit masked regions with structural guidance such as optical flow, depth, or keypoints, achieving temporal coherence but usually at the cost of losing fine-grained details. 
Recently, reference-guided methods~\citep{labs2025flux,deng2025emerging,wu2025qwen,hurst2024gpt,comanici2025gemini} have shown strong potential by leveraging reference images to combine flexible editing with high-fidelity reconstruction. 
Nonetheless, extending this paradigm to long video sequences remains challenging due to the scarcity of paired data and the difficulty of maintaining consistent identity or attributes over time.
In this work, we use reference-guided generation for face video swapping, enabling controllable identity transfer while preserving temporal coherence and visual fidelity across long sequences.

\section{Preliminary: Video Generation with DiT and Rectified Flow}

\label{sec:pre}






Recent advancements in diffusion-based video generation leverage the Diffusion Transformer (DiT) architecture combined with continuous-time training objectives such as Rectified Flow (RF)~\citep{esser2024scaling} to achieve high-quality and temporally coherent synthesis. 
DiT extends traditional UNet-based diffusion backbones with transformer blocks, enabling more flexible and scalable modeling of high-dimensional video data.
In this framework, the model learns a continuous denoising process by predicting the velocity between a pair of latent points. 
Given a ground-truth sample \( x_1 \), and a standard Gaussian noise \( x_0 \sim \mathcal{N}(0, I) \), a linearly interpolated latent \( x_t \) is constructed as:
\begin{equation}
x_t = t x_1 + (1 - t) x_0,
\end{equation}
where \( t \in [0, 1] \) is a timestep sampled from a predefined distribution. 
The target velocity is defined as the derivative of \( x_t \) with respect to time, yielding:
\begin{equation}
v_t = \frac{d x_t}{dt} = x_1 - x_0.
\end{equation}
The DiT model is trained to estimate this velocity given the latent \( x_t \), the conditioning signal \( c \), and the timestep \( t \). 
Let \( u(x_t, c, t; \theta) \) be the model's predicted velocity, where \( \theta \) denotes the model parameters. The training objective is to minimize the mean squared error (MSE) between the predicted and ground-truth velocities:
\begin{equation}
\mathcal{L} = \mathbb{E}_{x_0, x_1, c, t} \left\| u(x_t, c, t; \theta) - v_t \right\|^2.
\end{equation}
This training formulation enables high-quality results with significantly fewer steps and greater computational efficiency in video generation.



\section{Method}

\begin{figure*}[t]
        \centering
        \includegraphics[width=0.859\linewidth]{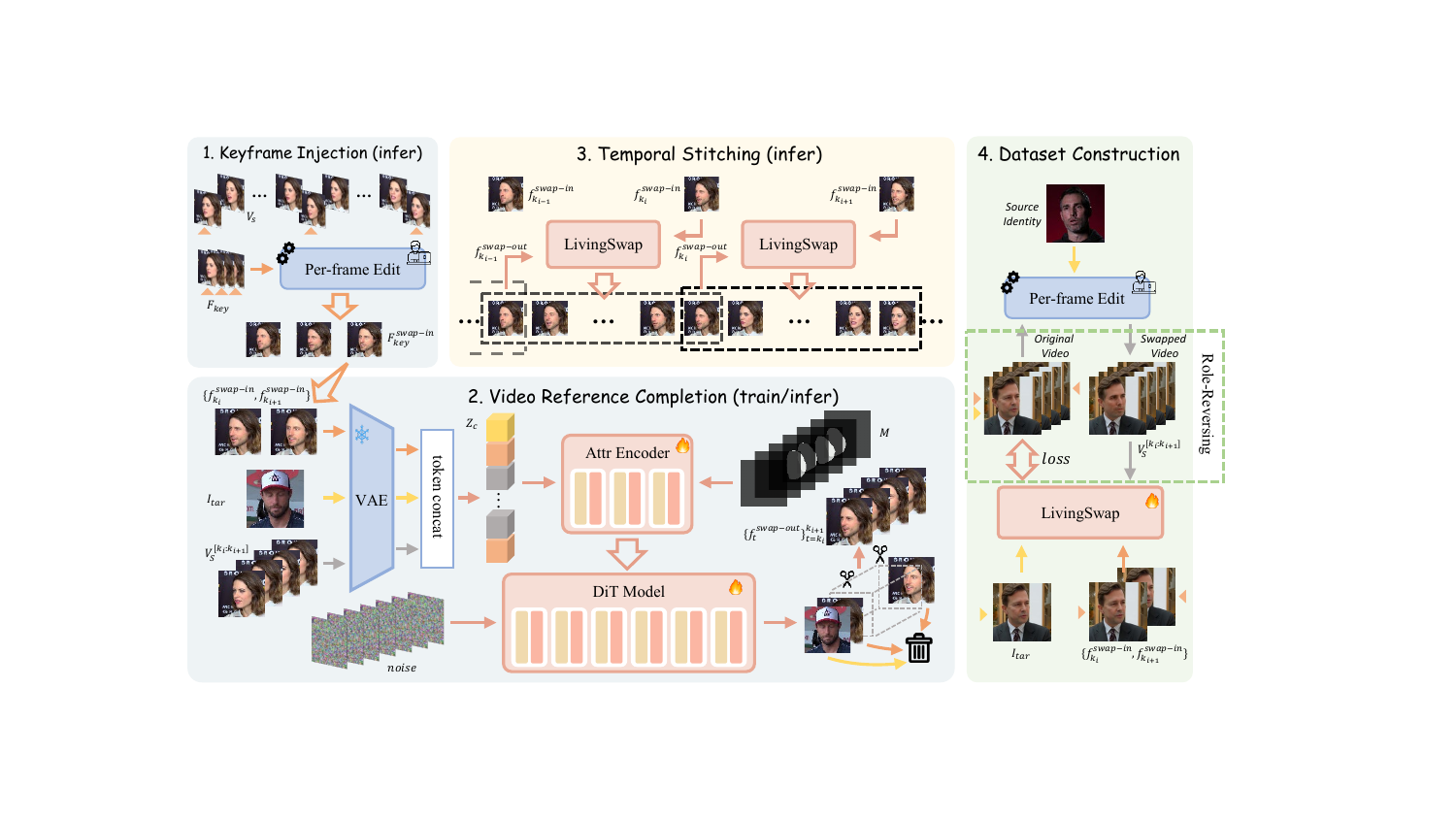}
        \vspace{-4pt}
        \caption{ 
        Overview of the proposed \modelname framework for video face swapping. 
        (1) Keyframes are used as temporal anchors to ensure consistent identity injection across long sequences. 
        (2) We feed the source video as a reference, enabling high-fidelity reconstruction of non-identity attributes such as lighting and expressions. 
        (3) By sequentially generating chunks and propagating the final frame of the previous chunk as guidance, \modelname achieves seamless transitions in long videos. 
        (4) We use Per-frame Edit method to generate the data and reverse data roles to construct paired samples, ensuring reliable and artifact-free learning.
        }
        \label{fig:pipeline}
\end{figure*}




In video face swapping tasks, the input typically consists of a source video 
$V_s = \{ f_t \mid t \in [1,T] \}$ to be modified, a mask sequence 
$M = \{ m_t \mid t \in [1,T] \}$ specifying the target regions for editing, 
and a target identity image $I_{\text{tar}}$.
The overall framework of \modelname is illustrated in \cref{fig:pipeline}.


In the following sections, we describe the design of \modelname across 
four essential components of video face swapping: 
(1) target identity injection, 
(2) preservation of source-video attributes, 
(3) consistent long-video generation, and 
(4) construction of a paired training dataset.
\subsection{Keyframes Identity Injection}

\label{sec:keyframe}





Maintaining a stable target identity across long and dynamic video sequences remains a fundamental challenge in video face swapping.
Compared to video-level approaches, image-level face swapping methods often provide stronger and more precise identity injection. 
Motivated by the complementary strengths of image-based face swapping and video interpolation paradigms~\citep{wang2024framer,guo2025keyframe}, we introduce a keyframe-based identity injection strategy that delivers robust and consistent identity conditioning for long-sequence video generation.

As illustrated in \cref{fig:pipeline} Part~1, we begin by selecting a set of representative frames from the input video as keyframes, denoted as
\begin{equation}
F_{\text{key}}^{\text{swap-in}} = { f_{k_i}^{\text{swap-in}} \mid k_i \in \mathcal{K},\ \mathcal{K} \subset [1, T] }.
\end{equation}
These keyframes are chosen at moments exhibiting significant variations in pose, expression, or illumination, ensuring that the major appearance changes across the video are well captured.
Next, we perform Per-frame Edit on these keyframes using a high-quality image-level method~\citep{facefusion2025}.
This process can be optionally followed by manual refinement using tools such as \textit{Adobe Photoshop}, which aligns with industrial workflows requiring both high-quality results and flexible editability.

After obtaining high-quality edited keyframes, we use each neighboring keyframe pair
\begin{equation}
\{ f_{k_i}^{\text{swap-in}},\; f_{k_{i+1}}^{\text{swap-in}} \}
\end{equation}
as a temporal boundary condition that guides the diffusion model during sequence completion.
Compared to per-frame processing commonly adopted in industrial pipelines, our keyframe guidance scheme requires modifying only a small set of boundary keyframes, which preserves the visual quality and flexibility of frame-by-frame editing while substantially improving efficiency.





\subsection{Video Reference Completion}

\label{sec:reference}

Beyond injecting the target identity, it is equally crucial to preserve non-identity attributes within the edited region and maintain the integrity of unmodified content in the source video. However, existing inpainting-based approaches typically discard original pixels and rely on additional structured inputs, which often degrade visual fidelity and weaken the generation prior.

Inspired by the reference-guided architecture of the video editing foundation model VACE~\cite{jiang2025vace}, we extend this paradigm to video face swapping to achieve high-fidelity reconstruction. As illustrated in \cref{fig:pipeline} Part~2, instead of masking the facial area in the source video and depending on external pretrained encoders, we directly feed the complete source video segment
\begin{equation}
V_{s}^{[k_i:k_{i+1}]} = \{ f_t \mid t \in [k_i, k_{i+1}] \}
\end{equation}
as a visual reference. This design preserves fine-grained visual cues—such as illumination, subtle expressions, and background details—without information degradation.

Furthermore, we incorporate an optional target identity image $I_{\text{tar}}$ to enhance identity fidelity in the first or last frame, particularly when the source video contains occlusions or low-quality instances (e.g., closed eyes). 
As demonstrated in our ablation study (\cref{tab:ablation_study}), this additional identity cue is not strictly necessary for all scenarios, but it consistently improves identity in challenging cases.


To integrate identity and appearance signals, we encode each input using the VAE encoder $\mathcal{E}_{\phi}(\cdot)$ and concatenate the resulting latent tokens in a temporally aligned order~\cite{zhao2025diceptiongeneralistdiffusionmodel}:
\begin{equation}
\begin{aligned}
Z_c = \operatorname{Concat}_{\text{token}}\big(
&\, \mathcal{E}_{\phi}(I_{\text{tar}}),\;
   \mathcal{E}_{\phi}(f_{k_i}^{\text{swap-in}}), \\
&\, \mathcal{E}_{\phi}(V_{s}^{[k_i:k_{i+1}]}),\;
   \mathcal{E}_{\phi}(f_{k_{i+1}}^{\text{swap-in}})
\big).
\end{aligned}
\end{equation}
where $Z_c$ denotes the aggregated latent conditioning tokens. The ordering naturally aligns with the temporal modeling of video diffusion models, enabling the generator to leverage temporal priors. Additionally, we construct a binary mask sequence $M$ that marks the editable region via black-filled tokens and concatenate it with $Z_c$ along the channel dimension to provide explicit spatial localization.

To support adaptive feature injection, we introduce an attribute encoder composed of DiT blocks, mirroring the architecture of the diffusion backbone and initialized with matching pretrained weights~\citep{zhang2023adding,jiang2025vace}. At each layer, the output of the attribute encoder is injected into the corresponding layer of the backbone via element-wise addition, enabling hierarchical and fine-grained conditioning in latent space. Formally, the injection process is defined as:
\begin{equation}
X^{(l+1)} = \mathcal{D}^{(l)}_{\theta}\!\Big(
X^{(l)} + \mathcal{A}^{(h)}_{\psi}(Z_c^{(h)}, M)
\Big),
\end{equation}
where $X^{(l)}$ is the hidden representation of the $l$-th layer of the diffusion backbone $\mathcal{D}_{\theta}$, and $\mathcal{A}^{(h)}_{\psi}$ denotes the $h$-th block of the attribute encoder. This design preserves the pretrained generative prior while enabling flexible and adaptive conditioning, thereby effectively capturing pixel-level details from the source video.

\subsection{Temporal Stitching}

\label{sec:stitch}








To accommodate industrial video face swapping requirements on variable-length inputs and to address the generation-length limitation of existing video diffusion backbones, we partition long videos into multiple fixed-length chunks for sequential processing. 
A critical question that follows is how to properly divide these chunks. Through extensive experiments, we find that introducing an overlap between adjacent chunks is essential.
When each chunk is generated independently without temporal overlap, noticeable frame discontinuities and temporal jumps often emerge at the boundaries.

Fortunately, benefiting from the synergy between keyframe design and reference-guided generation, our method enables efficient temporal stitching across chunks, ensuring coherent transitions in the final output.

Specifically, as illustrated in \cref{fig:pipeline} Part~3, we process the chunks in chronological order. 
To mitigate the accumulation of cross-chunk errors, we adopt the following effective strategy. 
For the first chunk, both the start and end guidance frames are taken directly from the corresponding keyframes. 
For subsequent chunks, we use the final output frame of the previous chunk, $f_{k_i}^{\text{swap-out}}$, as the start-frame guidance, while the end-frame guidance remains the keyframe $f_{k_{i+1}}^{\text{swap-in}}$. 
Formally, the generation of each chunk is defined as:
\begin{equation}
\begin{aligned}
\{ f_{t}^{\text{swap-out}} \}_{t = k_i}^{k_{i+1}} =
\mathcal{D}_{\theta,\psi}\!\big(
f_{k_i}^{\text{swap-out}},\;
f_{k_{i+1}}^{\text{swap-in}},\;
V_s^{[k_i:k_{i+1}]},\;
I_{\text{tar}},\;
M
\big).
\end{aligned}
\end{equation}

In addition, to flexibly position keyframes under the constraint of a fixed inference length, we employ several auxiliary engineering techniques, including frame interpolation, temporal reverse playback, frame skipping, and multi-pass inference.
These strategies allow the model to adapt to diverse video rhythms and temporal structures.

Finally, given that the reference model introduced in \cref{sec:reference} produces a fixed 81-frame output per inference, while our method typically requires manual editing only for the first and last frames of each chunk, this design yields approximately a \textbf{40×} reduction in manual labor, making the approach highly practical for industrial deployment.

\subsection{Dataset Construction}

\label{sec:dataset}

Face video datasets typically contain only the videos of individuals and lack paired source–target samples required for video reference face swapping.
To enable effective training of \modelname, we construct a paired dataset, \dataset, by combining a Per-frame Edit procedure with a role-reversing strategy to form source–target pairs.

As discussed in \cref{sec:keyframe}, the Per-frame Edit process provides high-quality single-frame face-swapped results.
This industrial approach, often based on GAN-based face swapping models such as Inswapper~\citep{facefusion2025}, achieves strong fidelity to the source video by leveraging the full-pixel source frame as input.
However, such results struggle with realism and suffer from temporal inconsistencies and degraded visual quality, including artifacts and distortions (\cref{fig:ab_data}).

To overcome these challenges, as illustrated in \cref{fig:pipeline} Part~4, we reverse the data pair roles when constructing training samples:
the GAN-generated swapped video is used as the model input $V_s$, while the original unedited video provides the keyframe inputs $F_{\text{key}}^{\text{swap-in}}$, the target identity image $I_{\text{tar}}$, and the ground-truth supervision.
This design ensures that the reference frames and ground-truth frames share the same identity, providing artifact-free, high-quality, and temporally consistent supervision signals.

Benefiting from the prior knowledge inherited from the pretrained model and our role-reversing strategy, \modelname exhibits strong robustness to noisy training samples, effectively going beyond the limitations of training data quality, as further demonstrated in \cref{sec:ablation}.
Finally, leveraging \dataset, we apply the rectified flow loss (detailed in \cref{sec:pre}) to supervise \modelname in preserving source-video attributes with high fidelity.

\begin{figure*}[ht!]
    \centering
    \vspace{-20pt}
    \includegraphics[width=0.95\linewidth]{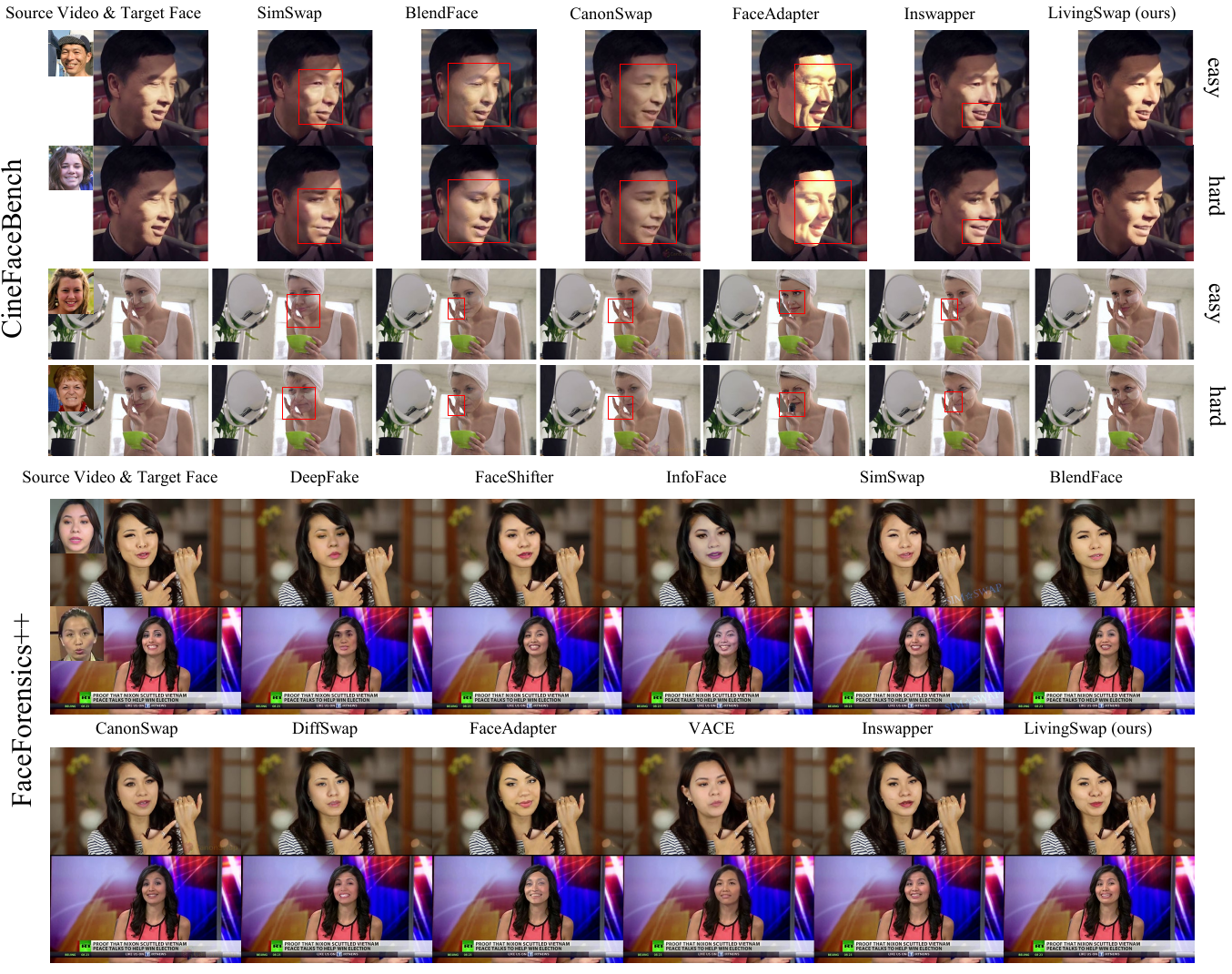}
    \caption{ 
    Qualitative comparison with state-of-the-art face-swapping methods.
    \modelname achieves the best overall performance, outperforming both GAN-based and diffusion-based approaches in video consistency, visual fidelity, and identity similarity.
    Although our keyframes are generated using Inswapper, the final results produced by \modelname are more stable and better preserve source attributes, even in challenging scenarios such as side profiles, occlusions, facial makeup, and complex lighting.
    }
    \label{fig:vis}
\end{figure*}

\section{Experiments}

\subsection{Experimental Setup}

\label{sec:setting}


\noindent
\textbf{Dataset.}
For training, we construct our dataset \dataset using CelebV-Text~\citep{yu2023celebv} and VFHQ~\citep{xie2022vfhq}.
CelebV-Text is a large-scale video–text dataset containing approximately 70,000 in-the-wild facial video clips, totaling around 279 hours of footage.
VFHQ comprises over 16,000 high-resolution facial video clips, covering diverse identities and scenarios.
Based on these two datasets, we employ Inswapper~\citep{facefusion2025} to generate paired face-swapping data and then reverse the Input-GroundTruth roles to construct our final training dataset, \textbf{\dataset}.
Please refer to \cref{sec:dataset} and the Supplementary Material for construction details.

\begin{table*}[t]
\centering

\caption{
Quantitative comparison \benchmark across multiple metrics.
Higher is better for ID Similarity and Gaze; lower is better for Expression, Lighting, Pose, and FVD. 
Best values are in \textbf{bold} and second best are \underline{underlined}.
}
\vspace{-5pt}

\resizebox{0.9005\textwidth}{!}{
\begin{tabular}{c|cc|cc|cc|cc|cc|cc|c}
\toprule
\multirow{2}{*}{Methods} 
& \multicolumn{2}{c|}{ID Sim.$\uparrow$} 
& \multicolumn{2}{c|}{Expr.$\downarrow$} 
& \multicolumn{2}{c|}{Light$\downarrow$} 
& \multicolumn{2}{c|}{Gaze$\uparrow$} 
& \multicolumn{2}{c|}{Pose$\downarrow$} 
& \multicolumn{2}{c|}{FVD$\downarrow$} 
& \multirow{2}{*}{Avg. Rank$\downarrow$} \\
& easy & hard 
& easy & hard
& easy & hard
& easy & hard
& easy & hard
& easy & hard
& \\
\midrule
SimSwap        
& 0.506 & \underline{0.385} 
& 2.217 & 2.683
& 0.214 & \underline{0.240}
& 0.722 & 0.712
& 4.623 & 4.820
& 74.63 & 75.33
& 3.917 \\
BlendFace      
& 0.482 & 0.315 
& \textbf{1.919} & \textbf{2.285}
& 0.245 & 0.271
& \underline{0.751} & 0.726
& 4.450 & 4.520
& 100.28 & 106.58
& 3.583 \\
CanonSwap      
& 0.506 & 0.365 
& \underline{1.935} & \underline{2.382}
& 0.223 & 0.255
& 0.671 & 0.684
& \underline{3.297} & \underline{3.492}
& 104.19 & 111.81
& 3.750 \\
Face-Adapter   
& 0.270 & 0.107 
& 2.208 & 2.495
& 0.291 & 0.319
& 0.685 & 0.692
& 5.643 & 6.423
& 176.96 & 182.25
& 5.583 \\
Inswapper      
& \textbf{0.567} & \textbf{0.422}
& 2.081 & 2.607
& \textbf{0.189} & 0.243
& 0.734 & \underline{0.741}
& 3.421 & 3.916
& \underline{66.62} & \underline{73.48}
& \underline{2.500} \\
LivingSwap (Ours)  
& \underline{0.532} & 0.367 
& 1.943 & 2.471
& \underline{0.192} & \textbf{0.238}
& \textbf{0.752} & \textbf{0.755}
& \textbf{3.108} & \textbf{3.399}
& \textbf{54.32} & \textbf{63.97}
& \textbf{1.667} \\
\bottomrule
\end{tabular}
}
\label{tab:quantitative_comparison_easy_hard}
\end{table*}

\begin{table}[t]
\centering

\vspace{-5pt}
\caption{Quantitative comparison with state-of-the-art 
methods on FF++. }

\vspace{-5pt}

\resizebox{0.48\textwidth}{!}{

\begin{tabular}{c|c|cccc|c|c}
\toprule
\multirow{1}{*}{Methods} 
& \multicolumn{1}{c|}{ID Sim. $\uparrow$} 
& Expr.$\downarrow$ 
& Light$\downarrow$ 
& Gaze$\uparrow$ 
& Pose$\downarrow$ 
& \multicolumn{1}{c|}{FVD$\downarrow$}  
& Avg. Rank$\downarrow$ \\ 
\midrule
Deepfakes      
& 0.432 
& 2.941 
& 0.340 
& 0.584 
& 4.662 
& 47.54 
& 9.50 \\
FaceShifter    
& 0.485 
& 2.451 
& 0.225 
& 0.690 
& 2.696 
& \textbf{18.73} 
& 4.67 \\
InfoSwap       
& 0.542 
& 2.868 
& 0.290 
& 0.586 
& 2.962 
& 47.28 
& 7.67 \\
SimSwap        
& 0.562 
& 2.674 
& 0.221 
& \textbf{0.720} 
& 2.977 
& 33.97 
& 5.17 \\
BlendFace      
& 0.480 
& \underline{2.256} 
& 0.228 
& \underline{0.717} 
& \underline{2.196} 
& 21.96 
& 4.00 \\
CanonSwap      
& 0.523 
& 2.307 
& \underline{0.205} 
& 0.685 
& \textbf{1.782} 
& 30.30 
& \underline{3.83} \\
DiffSwap       
& 0.261 
& \textbf{1.912} 
& \textbf{0.199} 
& 0.687 
& 2.277 
& 83.98 
& 5.00 \\
Face-Adapter   
& 0.247 
& 2.564 
& 0.259 
& 0.641 
& 3.608 
& 36.83 
& 8.17 \\
Inswapper      
& \textbf{0.636} 
& 2.536 
& 0.214 
& 0.704 
& 2.464 
& 20.63 
& \underline{3.83} \\
\midrule
LivingSwap (Ours)   
& \underline{0.592} 
& 2.466 
& 0.211 
& 0.706 
& 2.336 
& \underline{19.29} 
& \textbf{3.17} \\
\bottomrule
\end{tabular}
}
\label{tab:quantitative_comparison}

\vspace{-15pt}

\end{table}

\noindent
\textbf{Benchmark.}
For evaluation, we adopt FaceForensics++ (FF++)~\citep{rossler2019faceforensics++}, a widely used benchmark for video face manipulation analysis. 
However, FF++ primarily consists of interview-style or livestream videos, which do not adequately reflect the challenging conditions commonly encountered in cinematic productions.
To properly assess model performance in real film-like environments, we construct a new benchmark, \textbf{\benchmark}.
\benchmark contains 400 target–source test pairs. The curated video clips span a wide range of challenging cinematic scenarios, including long-take shots, complex lighting conditions, exaggerated expressions, heavy facial makeup, and semi-transparent occlusions.
To further assess robustness under different levels of identity similarity, each video is paired with two target images—an \textit{easy} and a \textit{hard} case—selected based on identity similarity scores.
Please refer to the Supplementary Material for construction details.




\noindent
\textbf{Metrics.}
To comprehensively evaluate the face-swapping performance, we employ both image-level and video-level metrics to assess the quality of the generated results.
Following prior work~\cite{chen2020simswap,chen2024hifivfs,wang2025dynamicface}, we randomly sample 10 frames from each face-swapped video to compute image-level evaluation metrics, including ID Similarity, Expression Error, Lighting Error, Gaze Error, and Face Pose Error.
ID Similarity is measured by encoding both the face-swapped result and the target image into identity vectors using a pre-trained ID encoder~\citep{wang2018cosface}, followed by computing the cosine similarity between them. 
In addition to identity-related metrics, we calculate Expression and Lighting Errors by extracting their respective coefficients using a 3DMM-based face reconstruction method~\citep{deng2019accurate} and computing the L2 distance between the source and swapped results. 
Similarly, we use a gaze estimation model~\citep{abdelrahman2023l2cs} with cosine similarity and a head pose estimation model~\citep{ruiz2018fine} with L2 distance to quantify the gaze and head pose changes.
For video-level evaluation, we use Frechet Video Distance (FVD)~\citep{unterthiner2018towards} to assess the overall quality of generated videos.
Additionally, we compute the average ranking of each method across all metrics to provide a comprehensive assessment of model performance.

\noindent
\textbf{Implementation Details.}
We initialize \modelname\ with the 14B pretrained weights from \citep{jiang2025vace}. 
The model is trained for 10,000 steps using the AdamW optimizer with a learning rate of 1e-5 and a batch size of 16. Following the original VACE configuration, the input resolution is set to 640 and the number of frames is set to 81. 
The final model is trained on 8 NVIDIA H200 GPUs for approximately 14 days.
During inference, we first detect faces using a pre-trained face detector, crop the detected regions, and perform face swapping on the cropped sequences. 
The swapped regions are then pasted back into their original locations in the frames.
To ensure reproducibility and fair comparison, we adopt a fixed-interval keyframe policy (every 79 frames for 81-frame chunks) in all experiments. 
As a face-swapping model widely used in industrial applications, we employ Inswapper~\citep{facefusion2025} as the per-frame editing method for processing keyframes, consistent with the \dataset construction.
For the ablation study, all models are trained for 2,000 steps with the same hyperparameters. 
In addition, we select the 100 most challenging samples from FF++ for evaluation to better highlight the effectiveness of each ablated variant.

\begin{figure*}[h]
    \centering
    \includegraphics[width=0.85\linewidth]{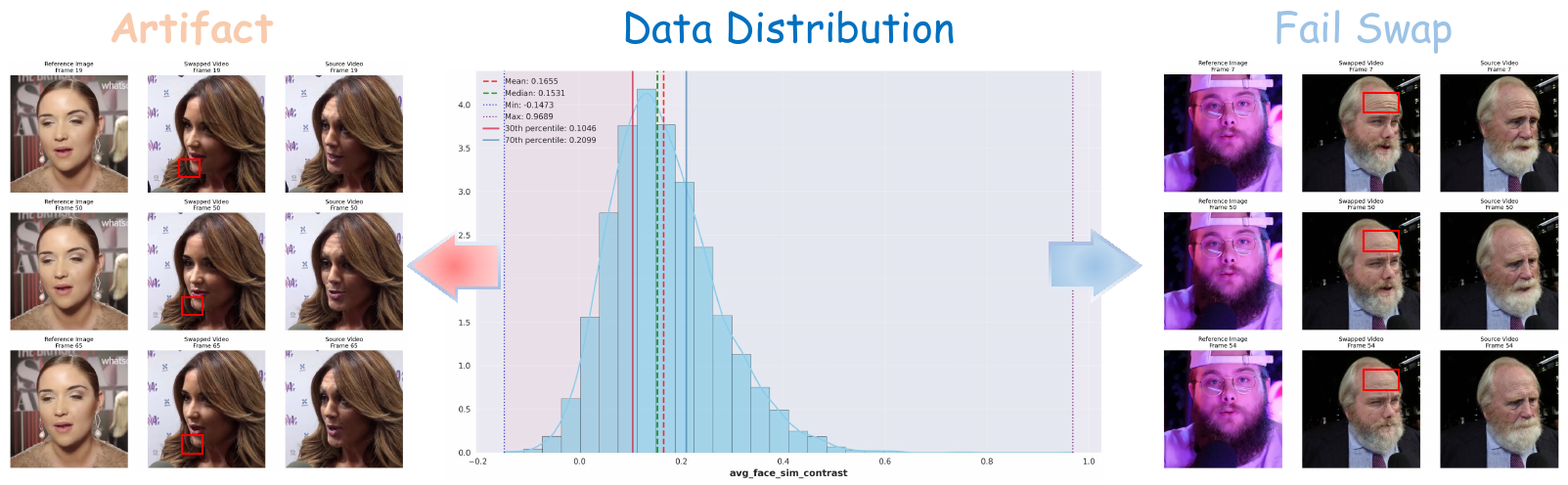}
    \vspace{-5pt}
    \caption{
    Visualization of the \textbf{\dataset} dataset. 
    The central plot shows the distribution of identity similarity scores between each swapped video and its corresponding original video, with the lowest 30\% (red) and highest 30\% (blue) highlighted. 
    Low-similarity pairs often contain artifacts and distortions as significant identity discrepancies (left), while high-similarity pairs may contain failed swap frames, causing identity inconsistencies and flickering (right).
    }
    \label{fig:ab_data}
    \vspace{-10pt}
\end{figure*}

\subsection{Comparisons with Existing Methods}

\label{sec:compare}

In this section, we compare our model against several state-of-the-art face-swapping approaches, including SimSwap~\citep{chen2020simswap}, InfoSwap~\citep{gao2021information}, BlendSwap~\citep{shiohara2023blendface}, CanonSwap~\citep{luo2025canonswap}, DiffSwap~\citep{zhao2023diffswap}, FaceAdapter~\citep{han2024face}, our baseline model VACE~\citep{jiang2025vace}, and the widely used industrial system Inswapper~\citep{facefusion2025}, which is also employed for our keyframe generation and dataset construction.
As shown in \cref{tab:quantitative_comparison}, on FF++—which mainly contains relatively simple scenarios such as interviews and livestreams—LivingSwap achieves the best overall ranking.

We further select the top-performing five methods on FF++ for additional evaluation on \benchmark.
As shown in \cref{tab:quantitative_comparison_easy_hard}, LivingSwap achieves state-of-the-art performance across multiple metrics and average ranks on both the \emph{easy} and \emph{hard} identity settings of our \benchmark, demonstrating strong robustness under varying identity similarity levels. 
Compared to our keyframe generation system, Inswapper, although our keyframes are generated from its outputs, our model exhibits superior temporal consistency, better preservation of source-video attributes, and more stable face-swapping behavior in difficult scenarios such as side views and occlusions (see \cref{fig:vis}). 
This also indicates that our approach is robust to imperfect or problematic keyframes, as further discussed in Supplementary Material.

\subsection{Ablation Studies}




\label{sec:ablation}

We conduct ablation studies on synthetic data quality, model design, keyframe quality, identity difference, and source video variation, with the results of the latter three experiments provided in the Supplementary Material.

\begin{table}[t]
\centering

\caption{
Ablation on generated data quality. Results show that using the full dataset achieves better fidelity (e.g., lighting, pose) due to greater sample diversity and demonstrate the model's robustness to failed noisy train data (also demonstrated in \cref{fig:ab_data2}).
}

\resizebox{0.45\textwidth}{!}{
\setlength{\tabcolsep}{1mm}
\begin{tabular}{c|c|cccc}
\toprule
\multirow{1}{*}{Methods} & \multicolumn{1}{c|}{ID Sim. $\uparrow$} & \multicolumn{1}{c|}{Expr.$\downarrow$} & \multicolumn{1}{c|}{Light$\downarrow$} & \multicolumn{1}{c|}{Gaze$\uparrow$} & \multicolumn{1}{c}{Pose$\downarrow$} \\ 
\midrule
LivingSwap          & 0.536 & 2.84 & \textbf{0.285} & 0.451 & \textbf{2.84} \\
VACE                   & 0.313 & 3.08 & 0.355 & 0.299 & 6.42 \\
Using Upper Data  & 0.532 & \textbf{2.82} & 0.289 & 0.484 & 2.89 \\
Using Lower Data  & \textbf{0.540} & 2.83 & 0.288 & \textbf{0.488} & 2.87 \\
\bottomrule
\end{tabular}
}

\vspace{-10pt}
\label{tab:ablation_data}
\end{table}

\noindent
\textbf{Ablation of Synthetic Data Quality.}
We investigate the impact of synthetic data quality using \dataset, constructed via Inswapper~\citep{facefusion2025}, which introduces flickering artifacts, identity inconsistencies, and various visual distortions (see \cref{fig:ab_data}). 
We first analyze the trade-off between quality and diversity by categorizing data into three groups based on identity variation (top 70\%, bottom 70\%, and full). 
Experimental results (\cref{tab:ablation_data}) reveal that our model is robust to data quality variations, as filtering yields no benefit. 
Instead, the diversity inherent in the full dataset proves critical for enhancing overall fidelity, leading us to adopt all available data.
To further verify whether \modelname is constrained by the quality of this training data, we evaluate it on manually selected noisy pairs containing artifacts. 
As illustrated in \cref{fig:ab_data2} in Supplementary Material, \textbf{\modelname consistently surpasses the quality of the original training samples}, producing results with improved expression alignment, visual realism, and significantly fewer local artifacts.
We attribute this ability to generalize beyond noisy supervision to two key designs: 
(1) our data reversing strategy, which ensures reliable ground-truth supervision even when input pairs are noisy, and 
(2) strong pretrained priors, which robustly correct corrupted patterns.

\begin{table}[t]
\centering

\caption{
Ablation of key components.
Our method achieves a more balanced trade-off between id preservation and fidelity.
Additionally, the w/o Target Image variant remains a viable option in scenarios where identity similarity is less critical.
}

\resizebox{0.45\textwidth}{!}{
\setlength{\tabcolsep}{1mm}
\begin{tabular}{c|c|cccc}
\toprule
\multirow{1}{*}{Methods} & \multicolumn{1}{c|}{ID Sim. $\uparrow$} & \multicolumn{1}{l}{Expr.$\downarrow$} & \multicolumn{1}{l}{Light$\downarrow$} & \multirow{1}{*}{Gaze$\uparrow$} & \multirow{1}{*}{Pose$\downarrow$}\\ 
\midrule
LivingSwap        & \textbf{0.536} & 2.84 & 0.285 & 0.451 & 2.84 \\
VACE                   & 0.313 & 3.08 & 0.355 & 0.299 & 6.42 \\
w/o Target Image  & 0.515 & 2.74 & 0.279 & \textbf{0.537} & \textbf{2.80} \\
w/o Keyframe & 0.281 & \textbf{2.47} & \textbf{0.249} & 0.502 & 2.84 \\
Inpainting   & 0.519 & 2.89 & 0.292 & 0.491 & 2.87 \\
\bottomrule
\end{tabular}
}

\vspace{-10pt}
\label{tab:ablation_study}
\end{table}




\noindent
\textbf{Ablation of Model Design.}
We conduct ablation studies on three key components of our model design: video reference, keyframe guidance, and target image reference. 
As shown in \cref{tab:ablation_study}, when we replace the video reference with the traditional inpainting approach, the model exhibits a decline in fidelity metrics. 
Regarding identity injection, when we remove the keyframe guidance and rely solely on the target image, we observe a significant drop in identity similarity as well as a noticeable degradation in temporal consistency. 
Conversely, when we ablate the identity provided by the target image, we still observe a decline in identity similarity. 
This is due to the limitations of keyframes in certain scenarios, such as occlusion, extreme angles, or closed eyes, which may result in the loss of critical identity features.


\section{Conclusion}




This work presented \modelname, the first video reference guided face swapping model that leverages keyframes as conditioning signals to enhance both fidelity and temporal coherence in video face swapping. 
By combining keyframe conditioning with video reference guidance, our approach ensures stable identity preservation and high-fidelity reconstruction across long video sequences. 
We propose a novel paired dataset, \dataset, along with a role-reversing strategy that provides reliable ground-truth supervision and tackles the challenge of scarce data for reference-guided training.
Meanwhile, we construct a new benchmark, \benchmark, for the video face swapping task in cinematic scenes.
Extensive experiments demonstrate that \modelname seamlessly integrates target identities with source video attributes while exhibiting strong performance.
Our model significantly reduces manual effort in production workflows, enabling more efficient and flexible video editing in film and entertainment.

\section*{Acknowledgments}

This work was supported by the National Natural Science Foundation of China (No. 62576315, No. 62506338).

{
    \small
    \bibliographystyle{ieeenat_fullname}
    \bibliography{main}
}

\clearpage
\setcounter{page}{1}
\maketitlesupplementary
\appendix


Considering the space constraints of the main paper, this supplementary material provides additional experimental results and presents the construction details of \dataset and \benchmark.
The content is organized as follows:
\begin{itemize}
    \item \cref{sec:com_data}: Generalization Beyond Train Data Quality.
    \item \cref{sec:im_id}: Keyframe Identity Injection for Accumulated Identity Errors.
    \item \cref{sec:ab_key}: Robustness to Keyframe Quality.
    \item \cref{sec:select_key}: Potential of Keyframe Selection.
    \item \cref{sec:ab_id}: Robustness to Identity Differences.
    \item \cref{sec:ab_source}: Robustness to Attribute Variations in Source Video.
    \item \cref{sec:grayscale}: Grayscale Keyframe Guidance for Robust Color Learning.
    \item \cref{sec:data_detail}: \dataset Construction Details.
    \item \cref{sec:bench_detail}: \benchmark Construction Details.
    \item \cref{sec:com_close}: Comparison with Closed-Source Methods.
    \item \cref{sec:limit}: Limitations of \modelname.
\end{itemize}

\begin{figure*}[h!]
    \centering
    \includegraphics[width=0.8\linewidth]{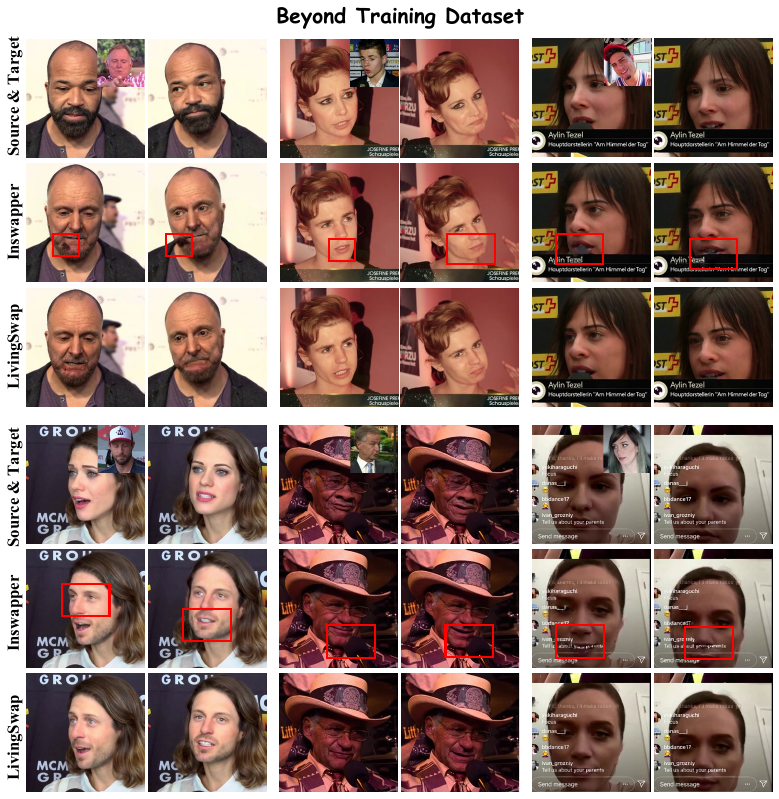}
    \caption{
    Qualitative comparison between the data pairs in \dataset (by Inswapper~\citep{facefusion2025}) and corresponding results generated by \modelname.
    Benefiting from reversing the role in data pair and strong priors in pretrained model, \modelname surpasses the quality of its training data, achieving better expression consistency and overall realism.
    Unlike Inswapper-based results, our method avoids local failure cases—such as incomplete swaps, mismatched regions, and occlusion-induced artifacts—demonstrating its strong generalization beyond the training dataset.
    }
    \label{fig:ab_data2}
\end{figure*}

\section{Generalization Beyond Train Data Quality} 

\label{sec:com_data}


In \cref{sec:ablation}, we analyzed the robustness of \modelname under varying levels of data quality, demonstrating the robustness of our model to failed noisy train data.
To further investigate whether \modelname is fundamentally constrained by the quality of the training dataset itself, we conduct an additional experiment.
We manually select several noisy source–target pairs from \dataset. 
These pairs contain various types of degradation, including local failure cases caused by failed swaps (e.g., residual beards), artifacts and misaligned expressions arising from large identity gaps or occlusions, as shown in \cref{fig:ab_data}. 
We then run \modelname on these noisy pairs using the exact same inference process as Inswapper, which was used to construct the dataset.

As illustrated in \cref{fig:ab_data2}, \modelname consistently surpasses the quality of the original dataset pairs, producing results with improved expression alignment, visual realism, and significantly fewer local artifacts.
We attribute this improvement to two key design choices.
(1) Reversing the role of data when constructing training pairs, which ensures reliable ground-truth supervision even when the original swaps contain noise.
(2) Strong priors in the pretrained model, which enable the system to robustly correct misaligned or corrupted supervision.
Together, these factors allow \modelname to generalize beyond the limitations of the training data and deliver high-quality, noise-resistant face swapping results.

\section{Keyframe Identity Injection for Accumulated Identity Errors} 

\label{sec:im_id}

\begin{figure*}[t]
    \centering
    \includegraphics[width=0.8\linewidth]{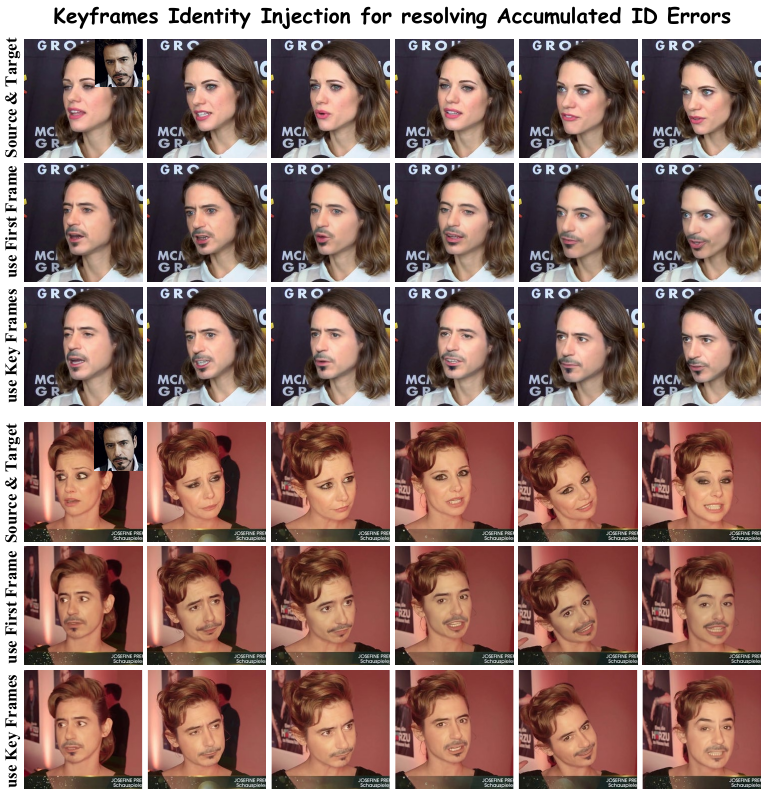}
    \caption{
    Keyframes Identity Injection for resolving Accumulated ID Errors.
    When using only the first frame for ID injection, face-swapping results suffer from gradually accumulating ID errors over time. 
    In contrast, with Keyframe Identity Injection, each video chunk is corrected individually by swapped keyframe, ensuring better ID consistency throughout the entire long video sequence.
    }
    \label{fig:im_id}
\end{figure*}


As demonstrated in \cref{sec:ablation}, keyframes play a crucial role in maintaining identity consistency within the video reference paradigm. 
The design of keyframes not only mitigates the interference caused by the source video’s identity but also plays a pivotal role in resolving accumulated ID errors. 
As shown in \cref{fig:im_id}, when using only the first frame as guidance and combining it with temporal stitching (as detailed in \cref{sec:stitch}) for long video generation, ID errors gradually accumulate as the video progresses, eventually leading to significant deviations from the target face.

In contrast, by injecting keyframe identities alongside temporal stitching, our method ensures a smooth connection with the previous video chunk while also providing correct ID guidance at the end of each chunk, preventing the accumulation of errors. 
This entire process is illustrated in the temporal stitching part of \cref{fig:pipeline}.

\section{Robustness to Keyframe Quality} 

\label{sec:ab_key}

\begin{figure*}[t]
    \centering
    \includegraphics[width=0.85\linewidth]{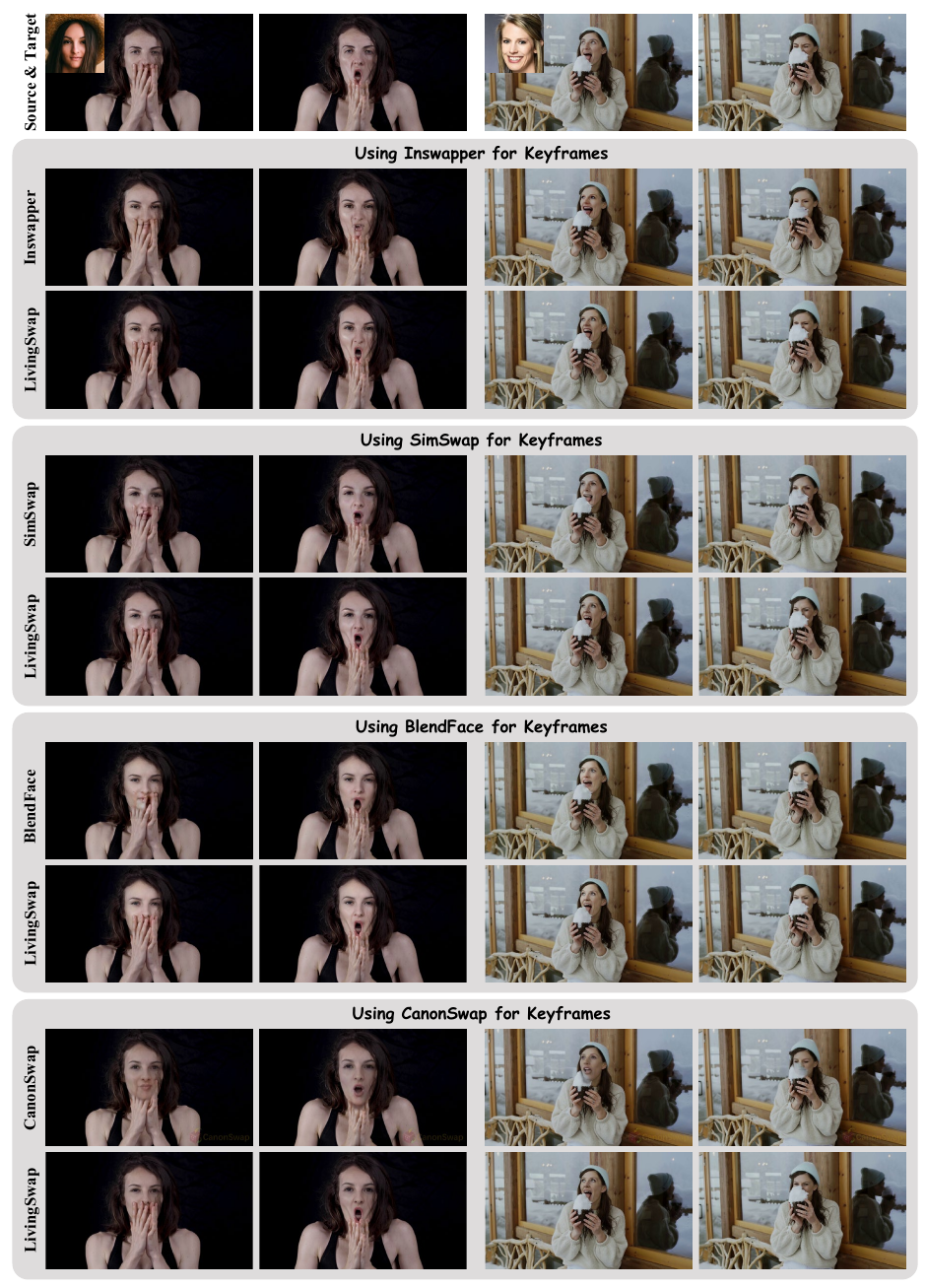}
    \caption{
    Qualitative comparison of using different image-level face swapping models as Per-frame Edit module.
    Injected keyframes often exhibit flaws including artifacts and expression misalignment.
    In contrast, by directly referencing the source video, \modelname successfully refines these flaws using the corresponding source attributes, demonstrating strong robustness to imperfect or corrupted keyframes.
    }
\label{fig:qualitative_keyframe_refinement}
\end{figure*}





To examine the sensitivity of \modelname to keyframe quality, we employ various image-level face swapping models as the Per-frame Edit module. 
As shown in ~\cref{fig:qualitative_keyframe_refinement}, \modelname demonstrates strong robustness against degraded or inconsistent keyframes. 
Even when the injected keyframes contain artifacts, expression misalignment, our method still produces results that remain well aligned with the source video and perceptually more faithful.
We attribute this robustness to two key factors:
(1) Directly referencing the source video, which enables the model to correct erroneous keyframe guidance by restoring the appropriate visual attributes; and
(2) The strong generative prior of diffusion models, which further enforces temporal realism and semantic consistency throughout the video.

\section{Potential of Keyframe Selection} 

\label{sec:select_key}

\begin{table}[t]
\centering
\caption{The potential of rule-based keyframe selection.}
\vspace{-10pt}

\resizebox{0.5\textwidth}{!}{
\begin{tabular}{c|ccccc}
\toprule
Methods & ID Sim. $\uparrow$ & Expr. $\downarrow$ & Light $\downarrow$ & Gaze $\uparrow$ & Pose $\downarrow$ \\ 
\midrule
Inswapper      
& 0.224 & 2.609 & 0.248 & 0.587 & 3.867 \\

LivingSwap (Fixed-interval Keyframe)
& 0.073 & 2.441 & 0.243 & 0.634 & \textbf{3.240} \\

LivingSwap (Rule-based Keyframe)
& \textbf{0.240} & \textbf{2.393} & \textbf{0.236} & \textbf{0.681} & 3.504 \\

\bottomrule
\end{tabular}
}

\vspace{-10pt}
\label{tab:keyframe_select}
\end{table}



Although \modelname demonstrates good robustness when handling keyframes with low face-swapping quality, using high-quality keyframes can significantly improve the overall video quality. 
Due to the limitations of current face-swapping models in handling profiles or extreme angles, we introduce a simple yet effective keyframe selection method: selecting frontal frames as keyframes based on face orientation detection (yaw within ±30° and pitch within ±20°). 
For evaluation, we selected the 10 worst-performing cases from the \benchmark for comparison. 
As shown in \cref{tab:keyframe_select}, using this straightforward keyframe selection rule greatly enhances the quality of the face-swapped video results.

Furthermore, by leveraging the keyframe guidance feature of \modelname, we can manually refine the face-swapping results or perform post-processing modifications (e.g., adjusting appearance or makeup) using tools like PhotoShop, providing greater flexibility in enhancing and fine-tuning the final face-swapping outcomes.

\section{Robustness to Identity Differences} 

\label{sec:ab_id}

\begin{figure*}[t]
    \centering
    \includegraphics[width=1.0\linewidth]{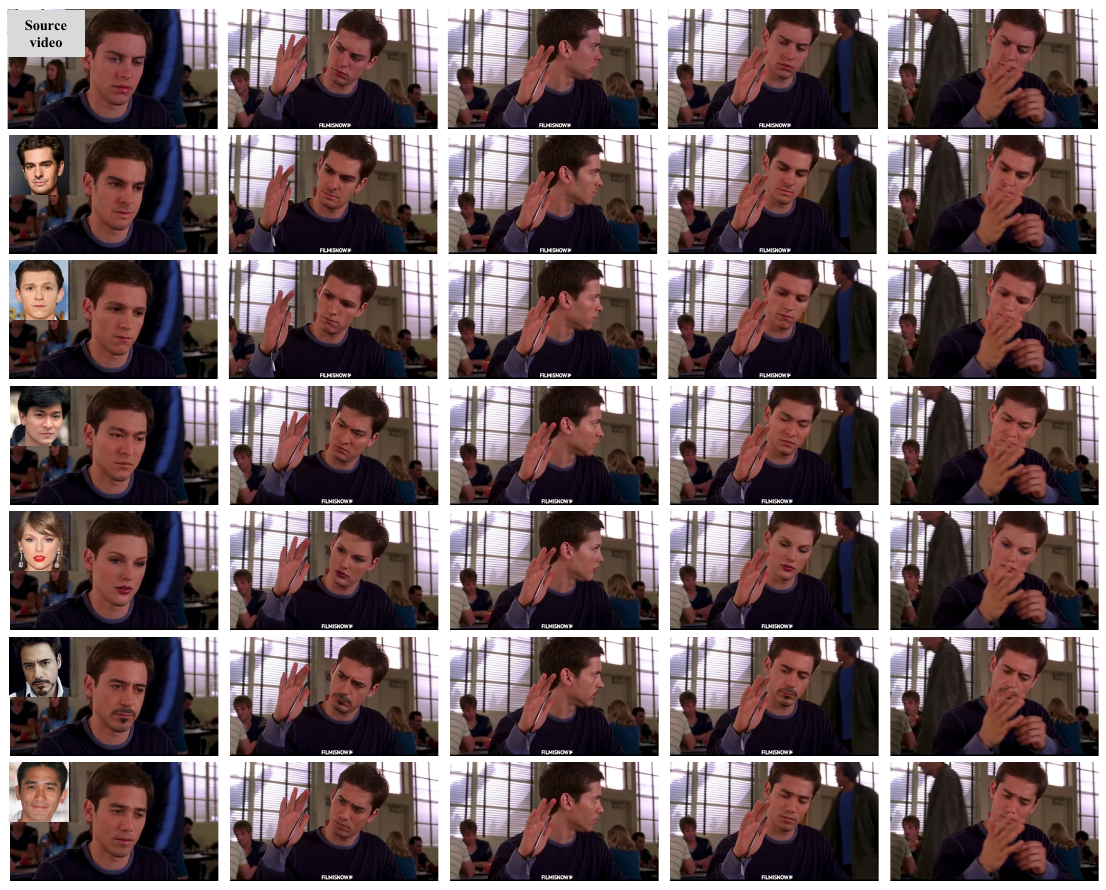}
    \caption{Identity swapping results on the same source video with different target identities. 
Our method produces consistent and high-fidelity face swaps regardless of large or small identity differences, demonstrating strong robustness to identity variations.}
    \label{fig:ab_id}
\end{figure*}


For the scenario of swapping different identities for the same source video, we conducted experiments with multiple videos and identities. 
As shown in \cref{fig:ab_id}, leveraging the advantages of keyframe identity injection, \modelname achieves satisfactory results for the same video, regardless of whether the identity difference is large or small. 
We hypothesize that this robustness of identity difference is due to the diversity of identities in our training data, as discussed in \cref{sec:ablation}.


\section{Robustness to Attribute Variations in Source Video} 

\label{sec:ab_source}

\begin{figure*}[t]
    \centering
    \includegraphics[width=1.0\linewidth]{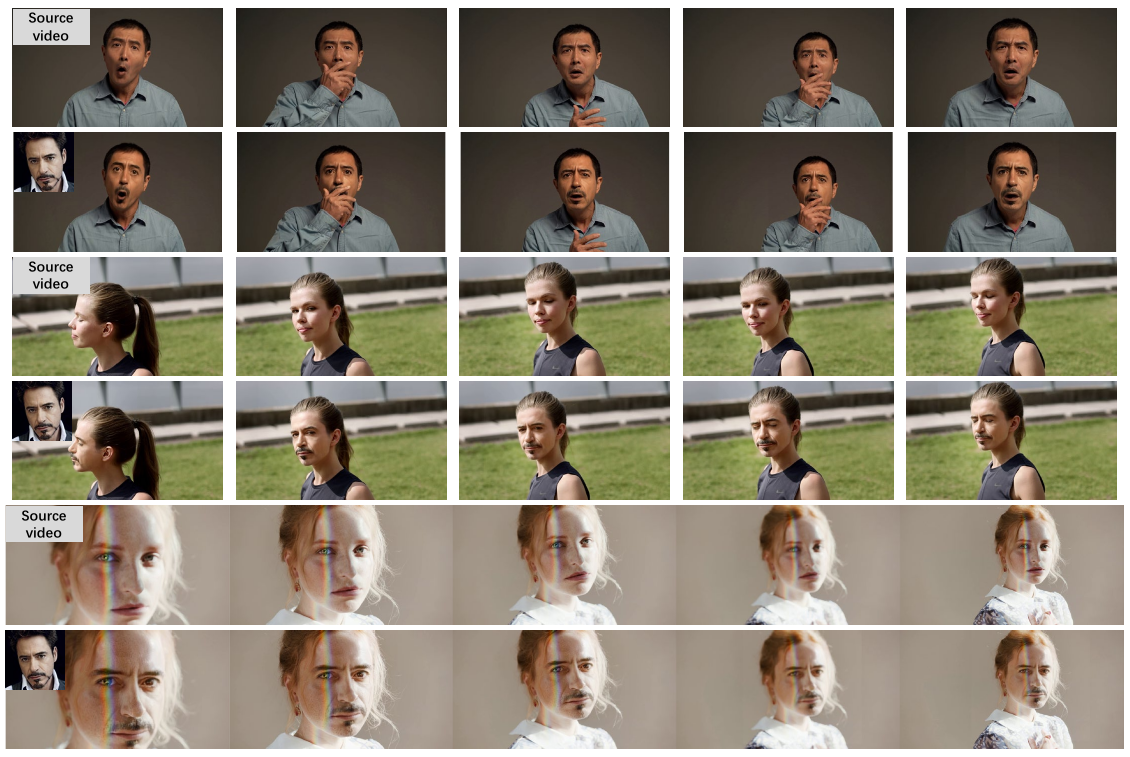}
    \caption{Face swapping results on diverse source videos with the same target identity. 
    Our method consistently preserves target identity and produces high-fidelity outputs across challenging conditions, including occlusions, side profiles, and complex lighting.}
    \label{fig:ab_src}
\end{figure*}


To verify whether our reference-based video face swapping approach is robust to attribute variations in the source video, we selected a diverse set of videos as source inputs and conducted experiments using the same target identity. 
As shown in \cref{fig:ab_src}, our model consistently produces high-quality results across attributes in challenging scenarios, such as occlusions, side profiles, and complex lighting conditions. 
Furthermore, owing to the robustness of keyframe quality, our model is able to generate realistic, high-fidelity outputs even when the keyframe model produces suboptimal results.





\section{Grayscale Keyframe Guidance for Robust Color Learning} 

\label{sec:grayscale}

\begin{figure*}[t]
    \centering
    \includegraphics[width=0.6\linewidth]{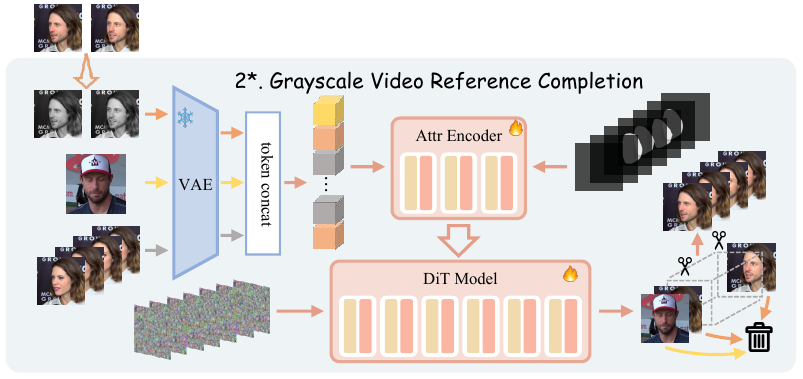}
    \caption{
    \textbf{Grayscale keyframe guidance.}
    To avoid incorrect color propagation from imperfect edited keyframes, we modify Video Reference Completion module and convert each keyframe to a grayscale image before VAE encoding. 
    This preserves structural cues (identity, pose, shading) while removing misleading chromatic information, allowing the model to recover accurate colors from the reference video.}
    \label{fig:grey_ref1}
\end{figure*}

\begin{figure*}[t]
    \centering
    \includegraphics[width=0.75\linewidth]{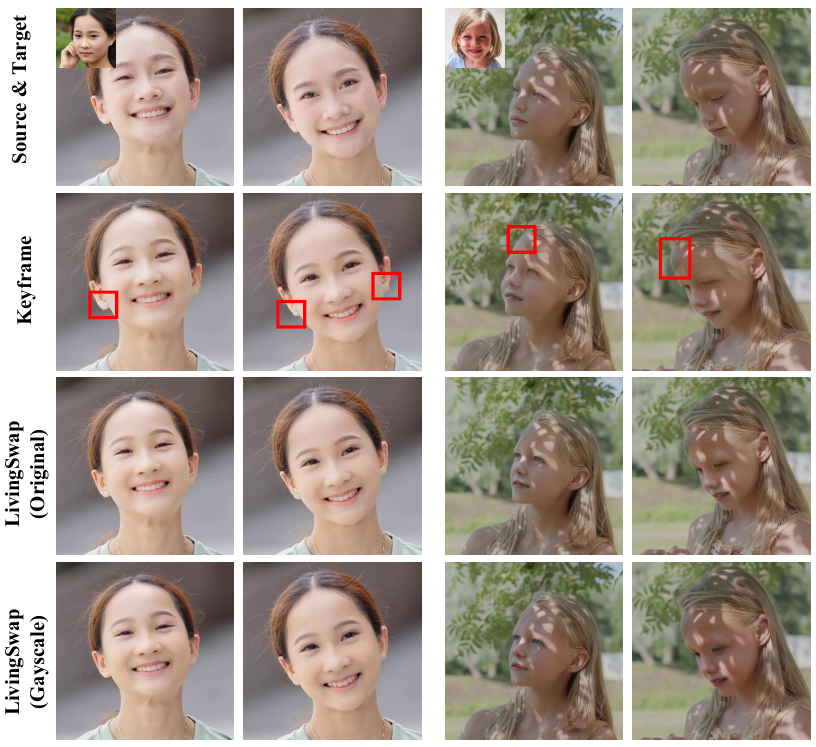}
    \caption{
    Compared with the original \modelname, using grayscale keyframes effectively suppresses color bleeding (e.g., the blue tint near the ear in the first example) and reduces temporal flickering artifacts (e.g., the dark patches on the head in the second example), leading to more stable and faithful video face swapping results.
    }
    \label{fig:grey_ref2}
\end{figure*}

As shown in \cref{fig:grey_ref2}, we observe that grayscale keyframe guidance significantly reduces color bleeding and flickering artifacts in challenging cases where the keyframe edits exhibit inconsistent or unrealistic colors, while maintaining comparable identity preservation and temporal consistency.

In the main paper, \modelname relies on RGB keyframes as temporal anchors for identity injection (\cref{fig:pipeline}). 
While this design is effective for propagating identity information, we observe a failure mode when the per-frame edited keyframes contain imperfect color statistics, e.g., incorrect skin tone or illumination caused by upstream editing tools, as shown in \cref{fig:grey_ref2}. 
Because the keyframe tokens are directly concatenated with the video tokens, such color biases can be mistakenly treated as a strong supervision signal, leading the diffusion model to reproduce the wrong colors in all synthesized frames.

To mitigate this issue, we introduce a simple yet effective modification: \emph{grayscale keyframe guidance}. 
As shown in \cref{fig:grey_ref1}, given an edited keyframe, we convert it to a single-channel luminance image, and then replicate this channel to form a three-channel grayscale keyframe before feeding it into the VAE encoder.
The rest of the pipeline, including token concatenation and the DiT-based video generation, remains unchanged.
This modification removes explicit chromatic information from the keyframe while preserving high-frequency structural cues such as facial identity, hairstyle, and local shading.

Intuitively, grayscale keyframes encourage the model to use the keyframe primarily as a structural and temporal anchor for stable identity injection, and to recover color statistics from the reference video branch in the Video Reference Completion module for fidelity.
As a result, LIVINGSWAP becomes less sensitive to color artifacts in the keyframe edits.

We finetune the final LivingSwap checkpoint for 5,000 steps to adapt it to grayscale pipeline.
As shown in \cref{fig:grey_ref2}, we observe that grayscale keyframe guidance significantly reduces color bleeding and flickering artifacts in challenging cases where the keyframe edits exhibit inconsistent or unrealistic colors, while maintaining comparable identity preservation and temporal consistency.

\section{\dataset Construction Details}

\label{sec:data_detail}



We construct our dataset \dataset based on CelebV-Text~\citep{yu2023celebv} and VFHQ~\citep{xie2022vfhq}.
First, we perform crop, resize, and clipping operations on the dataset to ensure the resolution is 640×640 pixels and the video length is approximately 200 frames. 
We then randomly pair the data and extract the first frame from the target video as the target face image. 
Next, we apply Inswapper~\citep{facefusion2025} to perform face-swapping on the entire dataset. 
The process is conducted using 8 NVIDIA H100 GPUs over a duration of 120 hours. 
Additionally, we use the face-parsing model~\citep{yu2018bisenet} to generate the face mask video. 
For the ablation study on the inpainting paradigm, we also use the pose estimation model~\citep{yang2023effective} to generate the corresponding pose video. 
After filtering out the failed samples from the processing steps, our dataset \dataset contains a total of 152,221 video samples, with a cumulative duration exceeding 300 hours.
Finally, we reverse the roles in each training pair: the swapped video is used as the source video (model input), while the original video serves as the target video (ground-truth supervision).

\section{\benchmark Construction Details}

\label{sec:bench_detail}

\begin{figure*}[t]
    \centering
    \includegraphics[width=1.0\linewidth]{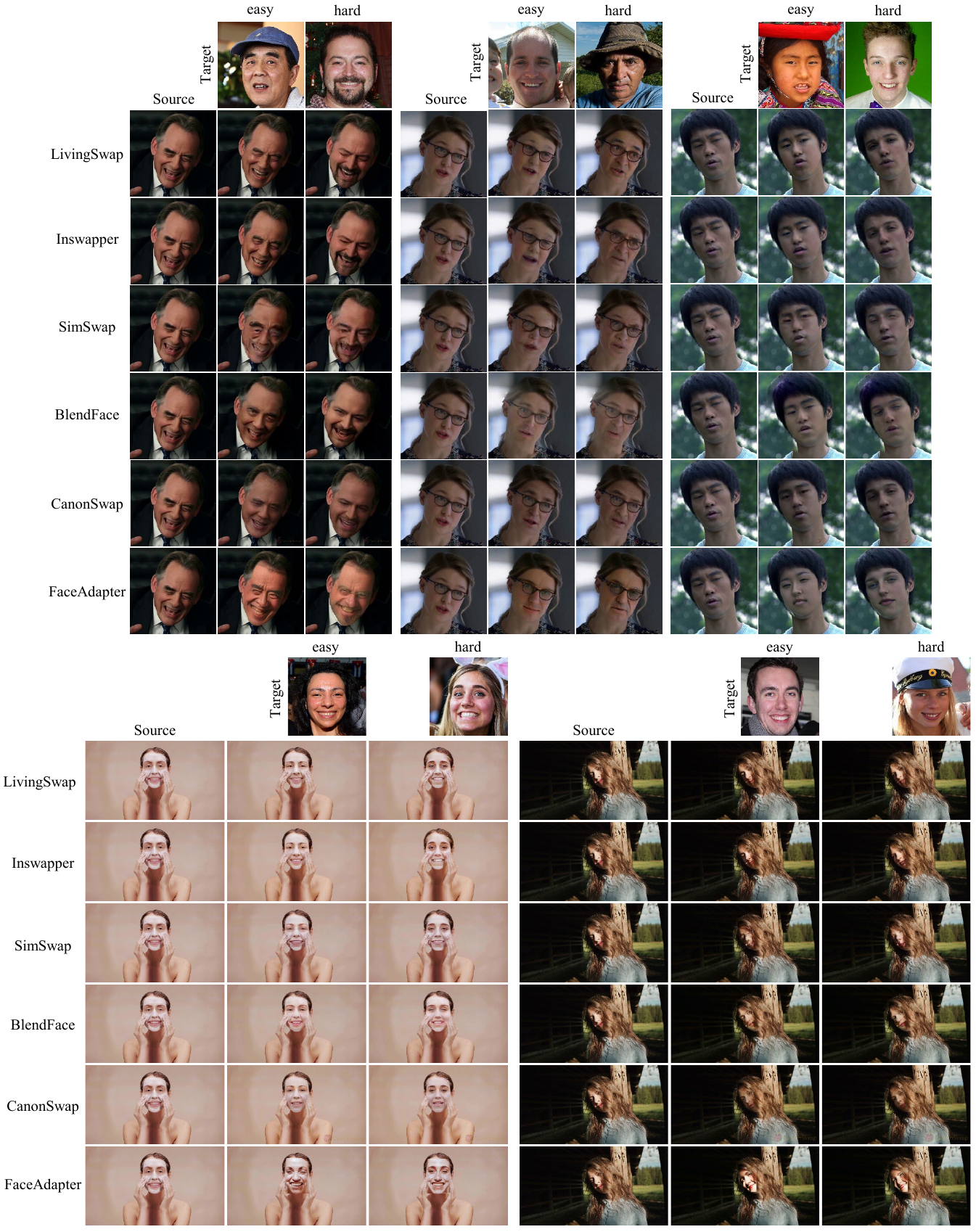}
    \caption{
    Additional Qualitative Comparison of Different Methods on \benchmark.
\modelname produces results with higher fidelity and realism compared to other methods.
    }
\label{fig:compare2}
\end{figure*}


Due to the fact that most scenes in FaceForensics++ \citep{rossler2019faceforensics++} consist of relatively simple settings such as interviews, hosts, or live broadcasts, it does not evaluate critical aspects often required in film scenarios, such as the model's ability to preserve facial expressions, lighting, makeup, and overall fidelity after face-swapping, as well as the stability of long video clips longer than 30 seconds. 
To address the limitations of the aforementioned benchmark, we have constructed a film scene face-swapping benchmark, \benchmark.

\benchmark consists of 200 video clips, paired with easy and hard target face images, resulting in 400 data pairs. 
Specifically, we downloaded and selected 100 video clips from two free video websites, Pixabay~\citep{pixabay} and Pexels~\citep{pexels}. 
Additionally, we selected 100 video clips from the OpenHumanVid dataset~\citep{li2025openhumanvid} and preprocessed them, resulting in 200 video samples used for evaluation. 
As shown in \cref{fig:compare2}, these 200 clips include challenging examples from film scenes, featuring difficult scenarios such as unique lighting, exaggerated expressions, micro-expressions, special makeup, occlusions, and even facial deformations.
In addition, there are several video cases that are longer than 1 minute.

On the other hand, we randomly selected 1,000 faces from the FFHQ dataset as target face images. 
By calculating the ID similarity (refer to \cref{sec:setting}) between these faces and the source video, we selected two samples with the most similar and least similar IDs to the source video, representing the easy and hard cases, respectively. 
This setup allows for a better evaluation of the model's robustness to ID differences. 
As shown in the \cref{tab:quantitative_comparison_easy_hard}, \cref{fig:vis} and \cref{fig:compare2}, our model demonstrates impressive performance in the challenging film scene scenarios.

\section{Comparison with Closed-Source Methods}

\label{sec:com_close}

\begin{figure*}[t]
    \centering
    \includegraphics[width=1.0\linewidth]{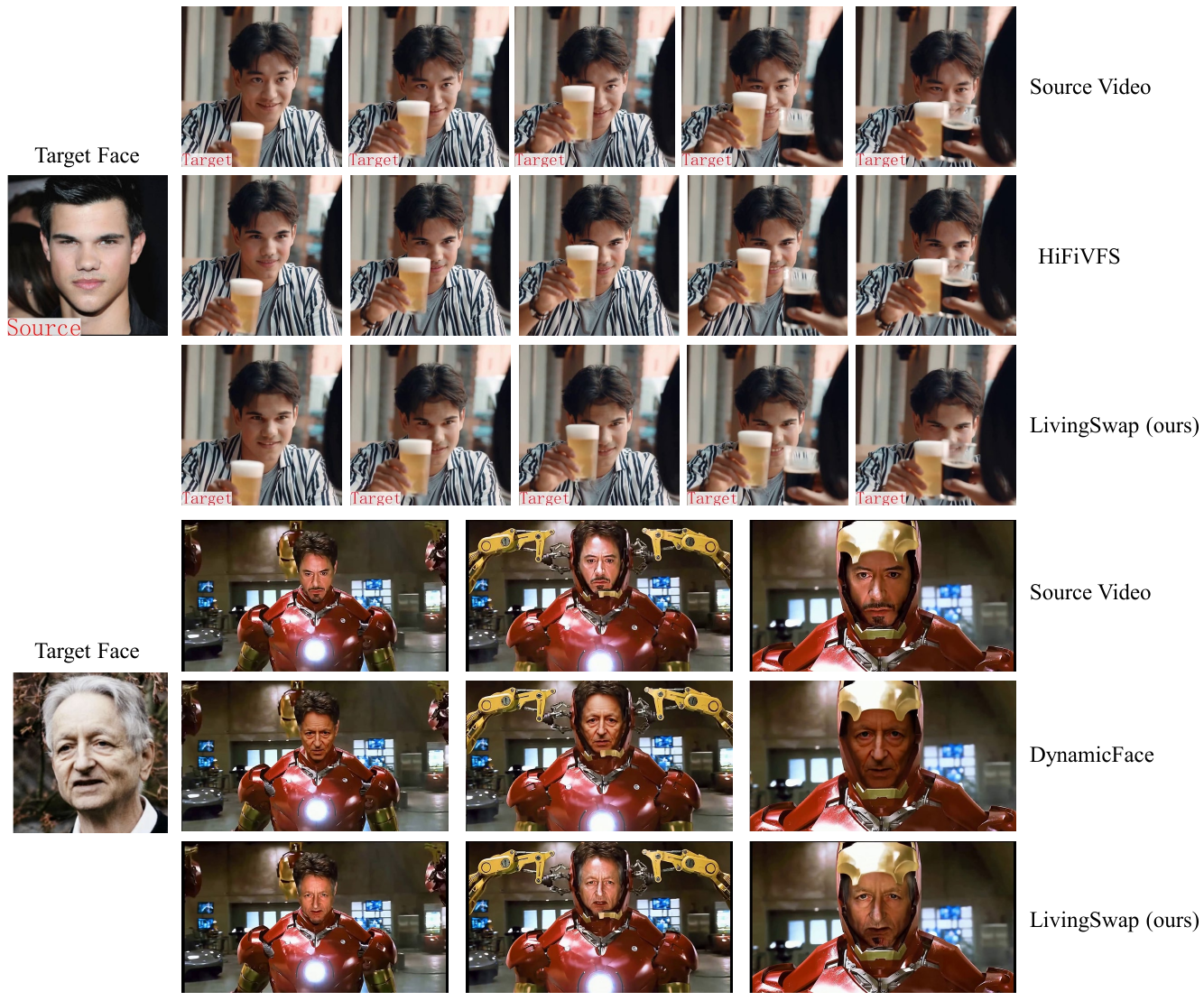}
    \caption{ 
    Qualitative comparison with recent inpainting-based video face swapping methods~\citep{chen2024hifivfs,han2024face} shows that our approach better preserves source video attributes (e.g., lighting and expression) and achieves greater stability under occlusions.
    }
    \label{fig:hifi}
\end{figure*}

Recently, several inpainting-based video face swapping methods using the Stable Video Diffusion model~\citep{blattmann2023stable} are proposed, such as HiFiVFS~\citep{chen2024hifivfs} and FaceAdapter~\citep{han2024face}. 
However, these methods are not open-source. 
To enable a comparison with them, we captured several demos from their project websites and conducted tests using the same target face image. 
The comparative results are shown in the \cref{fig:hifi}. 
Our approach better preserves the original video attributes such as lighting and expression, and also demonstrates strong stability in occluded cases.

\section{Limitations}

\label{sec:limit}



\modelname achieves better fidelity by directly referencing the source video, while enabling more flexible identity control and improved stability in long videos through keyframe identity injection. 
However, this framework design also introduces certain limitations:
1) Dependence on Keyframe Quality: 
The keyframe identity injection method creates a reliance on the quality of the selected keyframes. 
Although we demonstrated in \cref{sec:ab_key} that \modelname shows strong robustness to keyframe quality, it still encounters issues when dealing with keyframes that yield poor face-swapping results (e.g., identity or expression drift, image distortion, etc.). 
In such cases, the final output can be biased by the keyframe. 
This is supported by the experimental results in \cref{sec:select_key}, where selecting higher-quality keyframes led to significant improvements on the 10 worst-performing cases.
2) Slow Inference Speed: 
The use of a large video dataset to train the DiT model, combined with the Attribute Encoder for condition injection, significantly impacts the face-swapping speed of \modelname. 
In our experiments, for a video of 81 frames (approximately 3 seconds at 25 fps), \modelname requires 195 seconds (about 3 minutes) on a single H200 GPU with 108 GB memory. 
This translates to approximately 1 minute per second of swapped video.
In conclusion, our future work will focus on exploring better methods for identity injection and optimizing the face-swapping speed.

\end{document}